\documentclass[journal]{IEEEtran}

\usepackage[mathscr]{eucal}
\usepackage[cmex10]{amsmath}
\usepackage{epsfig,epsf,psfrag}
\usepackage{amssymb,amsmath,amsthm,amsfonts,latexsym}
\usepackage{amsmath,graphicx,bm,xcolor,url}
\usepackage[caption=false]{subfig} 
\usepackage{fixltx2e}
\usepackage{array}
\usepackage{verbatim}
\usepackage{bm}
\usepackage{algorithmic, cite}
\usepackage{algorithm}
\usepackage{verbatim}
\usepackage{textcomp}
\usepackage{mathrsfs}
\usepackage{epstopdf}

\catcode`~=11 \def\UrlSpecials{\do\~{\kern -.15em\lower .7ex\hbox{~}\kern .04em}} \catcode`~=13 

\allowdisplaybreaks[3]

\newcommand{\calC}{\mathcal{C}}

\newcommand{\calL}{\mathcal{L}}

\newcommand{\calN}{\mathcal{N}}
\newcommand{\calO}{\mathcal{O}}

\newcommand{\calT}{\mathcal{T}}

\newcommand{\ba}{\mathbf{a}}
\newcommand{\bA}{\mathbf{A}}

\newcommand{\bB}{\mathbf{B}}

\newcommand{\bC}{\mathbf{C}}
\newcommand{\bd}{\mathbf{d}}
\newcommand{\bD}{\mathbf{D}}
\newcommand{\be}{\mathbf{e}}
\newcommand{\bE}{\mathbf{E}}

\newcommand{\bF}{\mathbf{F}}

\newcommand{\bG}{\mathbf{G}}

\newcommand{\bI}{\mathbf{I}}

\newcommand{\bl}{\mathbf{l}}
\newcommand{\bL}{\mathbf{L}}

\newcommand{\bp}{\mathbf{p}}

\newcommand{\bQ}{\mathbf{Q}}

\newcommand{\bs}{\mathbf{s}}
\newcommand{\bS}{\mathbf{S}}

\newcommand{\bU}{\mathbf{U}}

\newcommand{\bV}{\mathbf{V}}

\newcommand{\bx}{\mathbf{x}}
\newcommand{\bX}{\mathbf{X}}

\newcommand{\bY}{\mathbf{Y}}
\newcommand{\bz}{\mathbf{z}}
\newcommand{\bZ}{\mathbf{Z}}

\DeclareMathAlphabet{\mathbsf}{OT1}{cmss}{bx}{n}
\DeclareMathAlphabet{\mathssf}{OT1}{cmss}{m}{sl}

\DeclareSymbolFont{bsfletters}{OT1}{cmss}{bx}{n}  
\DeclareSymbolFont{ssfletters}{OT1}{cmss}{m}{n}
\DeclareMathSymbol{\bsfGamma}{0}{bsfletters}{'000}
\DeclareMathSymbol{\ssfGamma}{0}{ssfletters}{'000}
\DeclareMathSymbol{\bsfDelta}{0}{bsfletters}{'001}
\DeclareMathSymbol{\ssfDelta}{0}{ssfletters}{'001}
\DeclareMathSymbol{\bsfTheta}{0}{bsfletters}{'002}
\DeclareMathSymbol{\ssfTheta}{0}{ssfletters}{'002}
\DeclareMathSymbol{\bsfLambda}{0}{bsfletters}{'003}
\DeclareMathSymbol{\ssfLambda}{0}{ssfletters}{'003}
\DeclareMathSymbol{\bsfXi}{0}{bsfletters}{'004}
\DeclareMathSymbol{\ssfXi}{0}{ssfletters}{'004}
\DeclareMathSymbol{\bsfPi}{0}{bsfletters}{'005}
\DeclareMathSymbol{\ssfPi}{0}{ssfletters}{'005}
\DeclareMathSymbol{\bsfSigma}{0}{bsfletters}{'006}
\DeclareMathSymbol{\ssfSigma}{0}{ssfletters}{'006}
\DeclareMathSymbol{\bsfUpsilon}{0}{bsfletters}{'007}
\DeclareMathSymbol{\ssfUpsilon}{0}{ssfletters}{'007}
\DeclareMathSymbol{\bsfPhi}{0}{bsfletters}{'010}
\DeclareMathSymbol{\ssfPhi}{0}{ssfletters}{'010}
\DeclareMathSymbol{\bsfPsi}{0}{bsfletters}{'011}
\DeclareMathSymbol{\ssfPsi}{0}{ssfletters}{'011}
\DeclareMathSymbol{\bsfOmega}{0}{bsfletters}{'012}
\DeclareMathSymbol{\ssfOmega}{0}{ssfletters}{'012}

\DeclareMathOperator{\sgn}{sgn}

\newtheorem{theorem}{Theorem} 
\newtheorem{lemma}[theorem]{Lemma}

\newtheorem{remark}{Remark}

\newtheorem{assumption}{Assumption}
\newtheorem{data model}{Data Model}

\newcommand{\qednew}{\nobreak \ifvmode \relax \else
      \ifdim\lastskip<1.5em \hskip-\lastskip
      \hskip1.5em plus0em minus0.5em \fi \nobreak
      \vrule height0.75em width0.5em depth0.25em\fi}

\usepackage{makecell}
\begin{document}
\title{Robust and Scalable Column/Row Sampling from Corrupted Big Data}

\author{Mostafa~Rahmani, \IEEEmembership{Student Member,~IEEE} and George~K.~Atia,~\IEEEmembership{Member,~IEEE} 
\thanks{This work was supported by NSF CAREER Award CCF-1552497 and NSF Grant CCF-1320547.

The authors are with the Department of Electrical and Computer Engineering, University of Central Florida, Orlando,
FL 32816 USA (e-mail: mostafa@knights.ucf.edu, george.atia@ucf.edu).}
}

\markboth{}%
{Shell \MakeLowercase{\textit{et al.}}: Bare Demo of IEEEtran.cls for Journals}
\maketitle

\begin{abstract}
Conventional sampling techniques fall short of drawing descriptive sketches of the data when the data is grossly corrupted as such corruptions break the low rank structure required for them to perform satisfactorily.
In this paper, we present new sampling algorithms which can locate the informative columns in presence of severe data corruptions. In addition, we develop new scalable randomized designs of the proposed algorithms.
The proposed approach is simultaneously robust to sparse corruption and outliers and substantially outperforms the state-of-the-art robust sampling algorithms as demonstrated by experiments conducted using both real and synthetic data.

\end{abstract}


\IEEEpeerreviewmaketitle

\section{Introduction}
Finding an informative or explanatory subset of a large number of data points is an important task of numerous machine learning and data analysis applications, including problems arising in computer vision \cite{elhamifar2014dissimilarity}, image processing \cite{esser2012convex}, bioinformatics \cite{bien2011prototype}, and recommender systems \cite{hartline2008optimal}. The compact representation provided by the informative data points helps summarize the data, understand the underlying interactions, save memory and enable remarkable computation speedups \cite{garcia2012prototype}.
Most existing sampling algorithms assume that the data points can be well approximated with low-dimensional subspaces. However,
much of the contemporary data comes with remarkable corruptions, outliers and missing values, wherefore a low-dimensional subspace (or a union of them) may not well fit the data. This fact calls for robust sampling algorithms, which can identify the informative data points in presence of all such imperfections. In this paper, we present an new column sampling approach which can identify the representative columns when the data is grossly corrupted and fraught with outliers.


\subsection{Summary of contributions}
We study the problem of informative column sampling in presence of sparse corruption and outliers. The key technical contributions of this paper are summarized next:
\textbf{I.} We present a new convex algorithm which locates the informative columns in  presence of sparse corruption with arbitrary magnitudes. \textbf{II.} We develop a set of randomized algorithms which provide scalable implementations of the proposed method for big data applications. We propose and implement a scalable column/row subspace pursuit algorithm 
that enables sampling in a particularly challenging scenario in which the data is highly structured.
\textbf{III.} We develop a new sampling algorithm that is robust to the simultaneous presence of sparse corruption and outlying data points. The proposed method is shown to outperform the-state-of-the-art robust (to outliers) sampling algorithms. \textbf{IV.} We propose an iterative solver for the proposed convex optimization problems.  


\subsection{Notations and data model}
 Given a matrix $\bL$, $\|\bL\|$ denotes its spectral norm, $\| \bL \|_1$ its $\ell_1$-norm given by $ \| \bL \|_1 = \sum\limits_{i,j} \big | \bL (i,j) \big |$, and
$\| \bL \|_{1,2}$ its $\ell_{1,2}$-norm defined as
$\| \bL \|_{1,2} = \sum\limits_{i}  \| \bl_i \|_2, $
where $\| \bl_i \|_2$ is the $\ell_2$-norm of the $i^{\text{th}}$ column of
$\bL$. In an $N$-dimensional space, $\be_i$ is the $i^{\text{th}}$ vector of
the standard basis. For a given vector $\ba$, $\| \ba \|_p$ denotes its $\ell_p$-norm. For a given matrix $\bA$, $\ba_i$ and $\ba^i$ are defined as the $i^{\text{th}}$ column and $i^{\text{th}}$ row of $\bA$, respectively.
In this paper, $\bL$ represents the low rank (LR) matrix (the clean data) with compact SVD $\bL = \bU \mathbf{\Sigma} \bV^T$, where $\bU \in \mathbb{R}^{N_1 \times r}$, $\mathbf{\Sigma} \in \mathbb{R}^{r \times r}$ and $\bV \in \mathbb{R}^{N_2 \times r}$ and $r$ is the rank of $\bL$.
Two linear subspaces $\calL_1$ and $\calL_2$ are independent if the dimension of their intersection $\calL_1 \cap \calL_2$ is equal to zero.
The incoherence condition for the row space of $\bL$ with parameter $\mu_v$ states that
 $ \underset{i}{\max} \: \| \be_i^T \bV \| \leq \mu_v r / N_2 $ \cite{lamport1}. 
In this paper (except for Section \ref{sec:bothcorruption}), it is assumed that the given data follows the following data model.

\begin{data model} The given data matrix $\bD \in \mathbb{R}^{N_1 \times N_2}$ can be expressed as $\bD = \bL + \bS$.   Matrix $\bS$ is  an element-wise sparse matrix with arbitrary support. Each element of $\bS$ is non-zero with a small probability $\rho$.
\end{data model}

\section{Related Work}
The vast majority of existing column sampling algorithms presume that the data lies in a low-dimensional subspace and look for few data points spanning the span of the dominant left singular vectors \cite{halko2011finding,tropp2009column}. The column sampling  methods based on the low rankness of the data can be generally categorized
into randomized \cite{deshpande2006matrix,drineas2006subspace,drineas2004clustering} and deterministic
methods \cite{boutsidis2008clustered,lashkari2007convex,esser2012convex,elhamifar2012see,gu1996efficient}. In the randomized method, the columns are sampled based on a carefully chosen probability distribution. For instance, \cite{drineas2004clustering} uses the $\ell_2$-norm of the columns, and in \cite{drineas2006subspace} the sampling probabilities are proportional to the norms of the
rows of the top right singular vectors of the data.
There are different types of deterministic sampling algorithms, including the rank revealing QR algorithm \cite{gu1996efficient}, and clustering-based algorithms \cite{boutsidis2008clustered}. In \cite{lashkari2007convex,esser2012convex,liu2015robustv,elhamifar2012see}
, sparse coding is used to leverage the self-expressiveness
property of the columns in low rank matrices to sample informative columns.

The low rankness of the data is a crucial requirement for these algorithms. For instance, \cite{drineas2006subspace} assumes that the span of few top right singular vectors approximates the row space of the data, and \cite{elhamifar2012see} presumes that the data columns admit sparse representations in the rest of the data, i.e., can be obtained through linear combinations of few columns. 
However, contemporary data comes with gross corruption and outliers. \cite{liu2015robustv} focused on column sampling in presence of outliers. While the approach in \cite{liu2015robustv} exhibits more robustness to the presence of outliers than older methods, it still ends up sampling from the outliers. In addition, it is not robust to other types of data corruption, especially element-wise sparse corruption. Element-wise sparse corruption can completely tear existing linear dependence between the columns, which is crucial for column sampling algorithms including \cite{liu2015robustv} to perform satisfactorily. 

\section{Shortcoming of Random Column Sampling}
\label{sec:randomsam}
As mentioned earlier, there is need for sampling algorithms capable of extracting important features and patterns in data when the data available is grossly corrupted and/or contains outliers.
Existing sampling algorithms are not robust to data corruptions, hence uniform random sampling is utilized to sample from corrupted data.
In this section, we discuss and study some of the shortcomings of random sampling in the context of two important machine learning problems, namely, data clustering and robust PCA.

\emph{Data clustering:}
Informative column sampling is an effective tool for data clustering \cite{lashkari2007convex}.  The representative columns are used as cluster centers and the data is clustered with respect to them. However, the columns sampled through random sampling may not be suitable for data clustering.
The first problem stems from the non-uniform distribution of the data points. For instance, if the population in one cluster is notably larger than the other clusters, random sampling may not acquire data points from the less populated clusters. The second problem is
that random sampling is data independent. Hence, even if a data point is sampled from a given cluster through random sampling, the sampled point may not necessarily be an important descriptive data point from that cluster.

\smallbreak
\emph{Robust PCA:}
There are many important applications in which the data follows Data model 1
  \cite{lamport1,lamport2,
lamport7,lamport16,minaee2016screennn,minaee2016screen,lamport19,lamport22}. In \cite{lamport1}, it was shown that the optimization problem
\begin{eqnarray}
\begin{aligned}
 \underset{\dot{\bL},\dot{\bS}}{\min} \: \lambda\|\dot{\bS}\|_1  + \|\dot{\bL} \|_* \quad
 \text{s. t.} \quad \dot{\bL} + \dot{\bS} = \bD \:
\end{aligned}
\label{eq2n}
\end{eqnarray}
is guaranteed to yield exact decomposition of $\bD$ into its LR and sparse components if the column space (CS) and the row space (RS) of $\bL$ are sufficiently incoherent with the standard basis.
However, the decomposition algorithms directly solving (\ref{eq2n}) are not scalable as they need to save the entire data in the working memory and have $\calO(r N_1 N_2)$ complexity per iteration.

An effective idea to develop scalable decomposition algorithms is to exploit the low dimensional structure of the LR matrix \cite{new2,rahmani2015high,rahmani2016subspace,new3,new1,liu2011solving}. The idea is to form a data sketch by sampling a subset of columns of $\bD$ whose LR component can span the CS of $\bL$. This sketch is decomposed using (\ref{eq2n}) to learn the CS of $\bL$. Similarly, the RS of $\bL$ is obtained by decomposing a subset of the rows. Finally, the LR matrix is recovered using the learned CS and RS. Thus, in lieu of decomposing the full scale data, one decomposes small sketches constructed from subsets of the data columns and rows.

Since existing sampling algorithms are not robust to sparse corruption, the scalable decomposition algorithms rely on uniform random sampling for column/row sampling. However, if the distributions of the columns/rows of $\bL$ are highly non-uniform, random sampling cannot yield \emph{concise} and descriptive sketches of the data.
For instance, suppose the columns of $\bL$ admit a subspace clustering structure \cite{rahmani2015innovation,elhamifar2013sparse} as per the following assumption.

\begin{assumption}
The matrix $\bL$ can be represented as $\bL = [\bU_1 \bQ_1 \: ... \: \bU_n \bQ_n]$. The CS of $\{ \bU_i \in \mathbb{R}^{N_1 \times r/n} \}_{i =1}^{n}$ are  random $r/n$-dimensional subspaces in $\mathbb{R}^{N_1}$. The RS of
$\{ \bQ_i \in \mathbb{R}^{r/n \times n_i} \}_{i =1}^{n}$ are random $r/n$-dimensional subspaces in $\{ \mathbb{R}^{n_i} \}_{i=1}^n$, respectively, $\sum_{i = 1}^n n_i = N_2$, and $\underset{i}{\min}\: {n_i} \gg r/n$.
\end{assumption}
%
The following two lemmas show that the sufficient number of randomly sampled columns to capture the CS can be quite large depending on the distribution of the columns of $\bL$.

\begin{lemma}
Suppose $m_1$ columns are sampled uniformly at random with replacement from the matrix $\bL$ with rank $r$. If
$
m_1 \ge 10 \mu_v r  \log \frac{2 r}{\delta},
$
then the selected columns of the matrix $\bL$ span the CS of $\bL$ with probability at least $1-\delta$.
\label{lm:smpl}
\end{lemma}

\begin{lemma}
If ~$\bL$ follows Assumption 1, the rank of $\bL$ is equal to $r$, $r/n \ge 18 \log \underset{i}{\max} \:\: n_i$ and $n_i \ge 96 \frac{r}{n} \log n_i, 1 \leq i \leq n $, then
$
\mathbb{P} \left[ \mu_v  <   \frac{1}{n} \frac{0.5 N_2}{\underset{i}{\min} \: n_i}  \right] \leq 2 \sum_{i = 1}^n n_i^{- 5}  \:.
$
\label{Lm:lowerbound}
\end{lemma}

\begin{figure}[h]
    \includegraphics[width=0.4\textwidth]{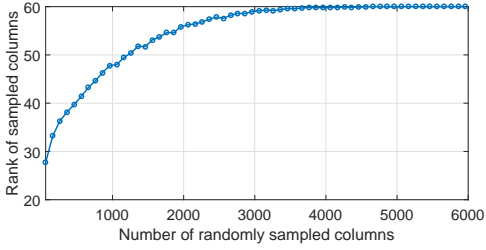}
    \centering
     \vspace{-0.01in}
    \caption{The rank of randomly sampled columns.}
    \label{fig:rand}
\end{figure}

\noindent
According to Lemma \ref{Lm:lowerbound}, the RS coherency parameter $\mu_v$ is linear in $\frac{N_2}{\underset{i}{min} \: n_i}$. The factor $\frac{N_2}{\underset{i}{min} \: n_i}$ can be quite large depending on the distribution of the columns. Thus, according to Lemma \ref{lm:smpl} and Lemma \ref{Lm:lowerbound}, we may need to sample too many columns to capture the CS if the distribution of the columns is highly non-uniform. As an example, consider $\bL = [\bL_1 \: \: \bL_2]$, the rank of $\bL = 60$ and $N_1=500$. The matrix $\bL_1$ follows Assumption 1 with $r=30$, $n = 30$ and $\{n_i\}_{i =1}^{30} = 5$. The matrix $\bL_2$ follows Assumption 1 with $r=30$, $n = 30$ and $\{n_i\}_{i =1}^{30} = 200$. Thus, the columns of $\bL \in \mathbb{R}^{500 \times 6150}$ lie in a union of 60 1-dimensional subspaces. Fig. \ref{fig:rand} shows the rank of randomly sampled columns of $\bL$ versus the number of sampled columns. Evidently, we need to sample
more than half of the data to span the CS. As such, we cannot evade high-dimensionality with uniform random column/row sampling.

\section{Column Sampling from Sparsely Corrupted Data}
In this section, the proposed robust sampling algorithm is presented. It is assumed that the data follows Data model 1. Consider the following optimization problem
\begin{eqnarray}
\underset{{\ba}}{\min} \| \bd_i - \bD_{-i} {\ba} \|_0 \quad \text{subject to}   \quad \| {\ba} \|_0 = r ,
\label{eq:l01n}
\end{eqnarray}
where $\bd_i$ is the $i^{\text{th}}$ column of $\bD$ and $\bD_{-i}$ is equal to $\bD$ with the $i^{\text{th}}$ column removed. If the CS of $\bL$ does not contain sparse vectors, the optimal point of (\ref{eq:l01n}) is equivalent to the optimal point of
\begin{eqnarray}
\begin{aligned}
& \underset{\ba}{\min}
& & \| \bs_i - \bS_{-i} {\ba} \|_0 \\
& \text{subject to}
& & \bl_i = \bL_{-i} \ba \quad \text{and} \quad \| \ba \|_0 = r \:,
\end{aligned}
\label{dokht}
\end{eqnarray}
where $\bl_i$ is the LR component of $\bd_i$ and $\bL_{-i}$ is the LR component of $\bD_{-i}$ ( similarly, $\bs_i$ and $\bS_{-i}$ are the sparse component).
To clarify, (\ref{eq:l01n}) samples $r$ columns of $\bD_{-i}$ whose LR component cancels out the LR component of $\bd_i$ and the linear combination $\bs_i - \bS_{-i} \ba$ is as sparse as possible.

\begin{figure}[h]
    \includegraphics[width=0.5\textwidth]{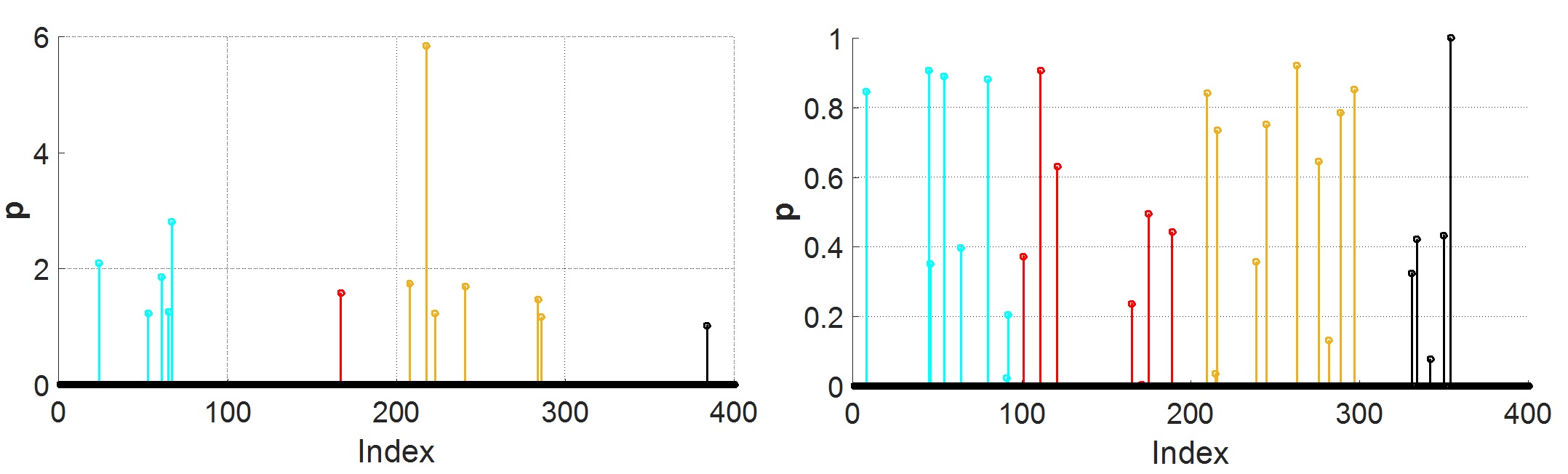}
    \centering
     \vspace{-0.17in}
    \caption{The elements of the vector $\bp$. Te left plot corresponds to (\ref{eq:mainopt}) and the right plot corresponds to \cite{elhamifar2012see}.}
    \label{fig:first}
\end{figure}

This idea can be extended by searching for a set of columns whose LR component can cancel out the LR component of all the columns. Thus, we modify (\ref{eq:l01n}) as
\begin{eqnarray}
\begin{aligned}
\underset{{\bA}}{\min} \| \bD - \bD {\bA} \|_0 \quad \text{s. t.}   \quad \| \bA^T \|_{0,2} = r,
\end{aligned}
\label{eq:firstid}
\end{eqnarray}
where $\| \bA^T \|_{0,2}$ is the number of non-zero rows of $\bA$.
The constraint in (\ref{eq:firstid}) forces $\bA$ to sample  $r$ columns. Both the objective function and the constraint in (\ref{eq:firstid}) are non-convex. We propose the following convex relaxation
\begin{eqnarray}
\begin{aligned}
\underset{{\bA}}{\min} \| \bD - \bD {\bA} \|_1  + \gamma \| \bA^T \|_{1,2} \:,
\end{aligned}
\label{eq:mainopt}
\end{eqnarray}
where $\gamma$ is a regularization parameter. Define $\bA^{*}$ as the optimal point of (\ref{eq:mainopt}) and define the vector $\bp \in \mathbb{R}^{N_2 \times 1}$ with entries $\bp(i) = \| {\ba^*}^i \|_2$, where $\bp(i)$ and ${\ba^*}^i$ are the $i^{\text{th}}$ element of $\bp$ and the $i^{\text{th}}$ row of ${\bA^*}$, respectively.
The non-zero elements of $\bp$ identify the representative columns.
For instance, suppose $\bD \in \mathbb{R}^{100 \times 400}$ follows Data model 1 with $\rho = 0.02$. The matrix $\bL$ with rank $r = 12$ can be expressed as $\bL = [ \bL_1 \: \: \bL_2 \: \: \bL_3 \: \: \bL_4]$ where the ranks of $\{ \bL_j \in \mathbb{R}^{100 \times 	100} \}_{j=1}^4$ are equal to 5, 1, 5, and 1, respectively. Fig. \ref{fig:first} shows the output of the proposed method and the algorithm presented in \cite{elhamifar2012see}. As shown, the proposed method samples a sufficient number of columns from each cluster. Since the algorithm presented in \cite{elhamifar2012see} requires strong linear dependence between the columns of $\bD$, the presence of the sparse corruption matrix $\bS$ seriously degrades its performance. 

\subsection{Robust column sampling from Big data}
The complexity of solving (\ref{eq:mainopt}) is $\calO(N_2^3 +N_1 N_2^2)$.
In this section, we present a randomized scalable approach which yields a scalable implementation of the proposed method for high dimensional data reducing complexity to $\calO(r^3 + r^2 N_2)$. We further assume that the RS of $\bL$ can be captured using a small random subset of the rows of $\bL$, an assumption that will be relaxed in Section \ref{sec:fully_structured}. The following lemma shows that the RS can be captured using few randomly sampled rows even if the distribution of the columns is highly non-uniform.
\begin{lemma}
Suppose $\bL$ follows Assumption 1 and $m_2$ rows of $\bL$ are sampled uniformly at random with replacement. If the rank of $\bL$
 is equal to $r$ and
\begin{eqnarray}
m_2 \ge 10 \: c \: r \varphi \log \frac{2 r}{\delta},
\label{eq:suff_row_com}
\end{eqnarray}
then the sampled rows span the row space of ~$\bL$ with probability at least $1 - \delta - 2 N_1^{-3}$, where $\varphi = \frac{\max (r ,  \log N_1 )}{r}$.
\label{lm:row_compare}
\end{lemma}
%
The sufficient number of sampled rows for the setup of Lemma \ref{lm:row_compare} is roughly $\calO(r)$, and is thus independent of the distribution of the columns. Let $\bD_r$ denote the matrix of randomly sampled rows of $\bD$ and $\bL_r$ its LR component. Suppose the rank of $\bL_r$ is equal to $r$. Since the RS of $\bL_r$ is equal to the RS of $\bL$, if a set of the columns of $\bL_r$ span its CS, the corresponding columns of $\bL$ will span the CS of $\bL$. Accordingly, we rewrite (\ref{eq:mainopt}) as
\begin{eqnarray}
\begin{aligned}
\underset{{\bA}}{\min} \| \bD_r - \bD_r {\bA} \|_1 ~+ \gamma \| \bA^T \|_{1,2} \:.
\end{aligned}
\label{eq:semif}
\end{eqnarray}
Note that we still have an $N_2 \times N_2$ dimensional optimization problem and the complexity of solving (\ref{eq:semif}) is roughly $\calO(N_2^3 + N_2^2 r)$.
In this section, we propose an iterative randomized method which solves (\ref{eq:semif}) with complexity $\calO(r^3 + N_2 r^2)$.
\textbf{Algorithm 1} presents the proposed solver. It starts the iteration with few randomly sampled columns of $\bD_r$ and refines the sampled columns in each iteration.

\begin{remark}
In both (\ref{eq:semif}) and (\ref{eq:D_s}), we use the same symbol $\bA$ to designate the optimization variable. However, in (\ref{eq:semif}) $\bA \in \mathbb{R}^{N_2 \times N_2}$, while in (\ref{eq:D_s}) $\bA \in \mathbb{R}^{m \times N_2}$, where $m$ of order $\calO(r)$ is the number of columns of $\bD_r^s$.  
\end{remark}

 \noindent
 Here we provide a brief explanation of the different steps of \textbf{Algorithm 1}:

\smallbreak
 \noindent
\textbf{Steps 2.1 and 2.2:} The matrix $\bD_r^s$ is the sampled columns of $\bD_r$. In steps 2.1 and 2.2, the redundant columns of $\bD_r^s$ are removed.

\smallbreak
\noindent
\textbf{Steps 2.3 and 2.4:} Define $\bL_r^s$ as the LR component of $\bD_r^s$. Steps 2.3 and 2.4 aim at finding the columns of $\bL_r$ which  do not lie in the CS of $\bL_r^s$. Define ${\bd_r}_i$ as the $i^{\text{th}}$ column of $\bD_r$. For a given $i$, if ${\bl_r}_i$ (the LR component of ${\bd_r}_i$) lies in the CS of $\bL_r^s$, the  $i^{\text{th}}$ column of $\bF = \bD_r - \bD_r^s \bA^{*}$ will be a sparse vector. Thus, if we remove a small portion of the elements of the $i^{\text{th}}$ column of $\bF$ with the largest magnitudes, the $i^{\text{th}}$ column of $\bF$ will approach the zero vector. Thus, by removing a small portion of the elements with largest magnitudes of each column of $\bF$, step 2.3 aims to locate the columns of $\bL_r$ that do not lie in the CS of $\bL_r^s$, namely, the columns of $\bL_r$ corresponding to the columns of $\bF$ with the largest $\ell_2$-norms. 
Therefore, in step 2.5, these columns are added to the matrix of sampled columns $\bD_r^s$.

\begin{figure}[h]
    \includegraphics[width=0.45\textwidth]{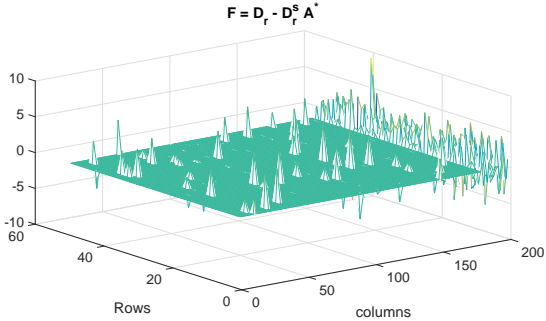}
    \centering
     \vspace{-0.02in}
    \caption{ Matrix $\bF$ in step 2.3 of Algorithm 1.  }
     \label{fig:explana}
 \end{figure}

As an example, suppose $\bD = [\bL_1 \: \: \bL_2] \,+\, \bS$ follows Data model 1 where the CS of $\bL_1 \in \mathbb{R}^{50 \times 180}$ is independent of the CS of $\bL_2 \in \mathbb{R}^{50 \times 20}$. In addition, assume $\bD_r = \bD$ and that all the columns of $\bD_r^s$ happen to be sampled from the first 180 columns of $\bD_r$, i.e., all sampled columns belong to $\bL_1$.  Thus, the LR component of the last 20 columns of $\bD_r$ do not lie in the CS of $\bL_r^s$. Fig. \ref{fig:explana} shows $\bF = \bD_r - \bD_r^s \bA^{*}$. One can observe that the algorithm will automatically sample the columns corresponding to $\bL_2$ because if few elements (with the largest absolute values) of each column are eliminated, only the last 20 columns will be non-zero.

\begin{remark}
The approach proposed in Algorithm 1 is not limited to the proposed method. The same idea can be used to enable more scalable versions of existing algorithms. For instance, the complexities of \cite{elhamifar2012see} and \cite{liu2015robustv} can be reduced from roughly $N_2^3$ to $N_2 r$, which is a remarkable speedup for high dimensional data.
\end{remark}

\begin{algorithm}
\caption{Scalable Randomized Solver for (\ref{eq:semif}) }
{\footnotesize
\textbf{1. Initialization}\\
\textbf{1.1} Set $c_1$, $c_2$, and $k_{\max}$ equal to integers greater than 0. Set $\tau$ equal to a positive integer less than 50. $\hat{r}$ is a known upper bound on $r$.

\textbf{1.2}  Form $\bD_r \in \mathbb{R}^{m_2 \times N_2}$  by sampling $m_2 = c_1 \hat{ r}$ rows of $\bD$ randomly.

\textbf{1.3} Form $\bD_{r}^s \in \mathbb{R}^{m_2 \times c_2\hat{r}}$ by sampling  $c_2 \hat{r}$ columns  of $\bD_r$  randomly.

\smallbreak
\textbf{2. For} $k$ from 1 to $k_{\max}$
\\
 \textbf{2.1 Locate informative columns:} Define $\bA^{*}$ as the optimal point of
 \begin{eqnarray}
\underset{{\bA}}{\min} \| \bD_r - \bD_r^s \bA \|_1 + \gamma \| \bA^T \|_{1,2} \:.
\label{eq:D_s}
\end{eqnarray}

\textbf{2.2 Remove redundant columns:}
Remove the zero rows of $\bA^{*}$ (or the rows with $\ell_2$-norms close to 0) and remove the columns of $\bD_r^s$ corresponding to these zero rows.

\textbf{2.3 Remove sparse residuals:}  Define $\bF =  \bD_r - \bD_r^s \bA^{*}  $. For each column of $\bF$, remove the $\tau$ percent elements with largest absolute values.

\textbf{2.4 Locate new informative columns:}
 Define $\bD_r^n \in \mathbb{R}^{m_2 \times 2 \hat{r}}$ as the columns of $\bD_r$ which are corresponding to the columns of $\bF$ with maximum $\ell_2$-norms.

\textbf{2.5 Update sampled columns:} $\bD_r^s = [\bD_r^s \: \: \bD_r^n]$.

\textbf{2. End For}

\textbf{Output:} Construct $\bD_c$ as the columns of $\bD$ corresponding to the sampled columns from $\bD_r$ (which form $\bD_r^s$).  The columns of $\bD_c$ are the sampled columns.
}
\end{algorithm}

\subsubsection{Sampling from highly structured Big data}
\label{sec:fully_structured}
\textbf{Algorithm 1} presumes that the rows of $\bL$ are well distributed such that $c_1 \hat{r}$ randomly sampled rows of $\bL$ span its RS. This may be true in many settings where the clustering structure is only along one direction (either the rows or columns), in which case \textbf{Algorithm 1} can successfully locate the informative columns. \textcolor{black}{ If, however, both the columns and the rows exhibit clustering structures and their distribution is highly non-uniform, neither the CS nor the RS can be captured concisely using random sampling. As such, in this section we address the scenario in which the rank of $\bL_r$ may not be equal to $r$.} We present an iterative CS-RS pursuit approach which converges to the CS and RS of $\bL$ in few iterations.

\begin{algorithm}
\caption{  {Column/Row Subspace Pursuit Algorithm}}
{\footnotesize
\textbf{Initialization:} Set $\bD_w$ equal to $c_1 \hat{r}$ randomly sampled rows of $\bD$. Set $\bX$ equal to $c_2 \hat{r}$ randomly sampled columns of $\bD_w$ and set $k_{\max}$ equal to an integer greater than 0.

\smallbreak

\textbf{For} $j$ from 1 to $j_{\max}$ do

\smallbreak
\textbf{1. Column Sampling} \\
\textbf{1.1 Locating informative columns:} Apply Algorithm 1 without the initialization step to $\bD_w$ as follows: set $\bD_r$ equal to $\bD_w$, set $\bD_r^s$ equal to $\bX$, and set $k_{\max} = 1$. \\
\textbf{1.2 Update sub-matrix $\bX$:} Set sub-matrix $\bX$ equal to $\bD_r^s$, the output of Step 2.5 of Algorithm 1. \\
\textbf{1.3 Sample the columns:} Form matrix $\bD_c$ using the columns of $\bD$ corresponding to the columns of $\bD_w$ which were used to form $\bX$.

\smallbreak
\textbf{2. Row Sampling}\\
\textbf{2.1 Locating informative rows:} Apply Algorithm 1 without the initialization step to $\bD_c^T$ as follows: set $\bD_r$ equal to $\bD_c^T$, set $\bD_r^s$ equal to $\bX^T$, and set $k_{\max} = 1$.

\textbf{2.2 Update sub-matrix $\bX$:} Set sub-matrix $\bX^T$ equal to $\bD_r^s$, the output of step 2.5 of Algorithm 1.

\textbf{2.3 Sample the rows:} Form matrix $\bD_w$ using the rows of $\bD$ corresponding to the rows of $\bD_c$ which were used to form $\bX$.

\smallbreak
\textbf{End For}

\smallbreak
\textbf{Output:} The matrices $\bD_c$ and $\bD_w$ are the sampled columns and rows, respectively.
}
\end{algorithm}



\begin{figure}[h]
    \includegraphics[width=0.42\textwidth]{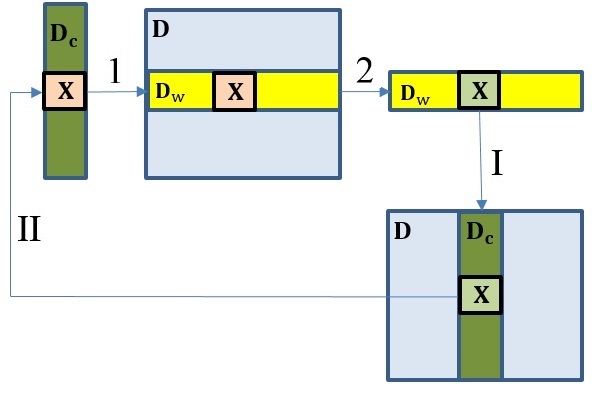}
    \centering
     \vspace{-0.05in}
    \caption{ Visualization of Algorithm 2. \textbf{Iّ}: Matrix $\bD_c$ is obtained as the columns of $\bD$ corresponding to the columns which form $\bX$. \textbf{IّI}: Algorithm 1 is applied to Matrix $\bD_c$ to update $\bX$ (the sampled rows of $\bD_c$). \textbf{1}: Matrix $\bD_w$ is obtained as the rows of $\bD$ corresponding to the rows which form $\bX$. \textbf{2}: Algorithm 1 is applied to matrix $\bD_w$ to update $\bX$ (the sampled columns of $\bD_w$). }
    \label{alg3}
\end{figure}

The table of \textbf{Algorithm 2}, Fig. \ref{alg3} and its caption provide the details of the proposed sampling approach along with the definitions of the used matrices.
 We start the cycle from the position marked I in Fig. \ref{alg3}.
The matrix $\bX$ is the informative columns of $\bD_w$. Thus, the rank of $\bX_L$ (the LR component of $\bX$) is equal to the rank of ${\bL}_w$ (the LR component of $\bD_w$). The rows of $\bX_L$ are a subset of the rows of $\bL_c$ (the LR component of $\bD_c$). If the rows of $\bL$ exhibit a clustering structure, it is likely that rank$(\bX_L)< \text{rank} (\bL_c)$. Thus, rank$({\bL}_w)< \text{rank} (\bL_c)$. 
We continue one cycle of the algorithm by going through steps  II and 1 of Fig. \ref{alg3} to update $\bD_w$. Using a similar argument, we see that the rank of an updated $\bL_w$ will be greater than the rank of $\bL_c$. Thus, if we run more cycles of the algorithm -- each time updating $\bD_w$ and $\bD_c$ -- the rank of $\bL_w$ and $\bL_c$ will increase.
 While there is no guarantee that the rank of $\bL_w$ will converge to $r$ (it can converge to a value smaller than $r$), our investigations have shown that \textbf{Algorithm 2} performs quite well and the RS of $\bL_w$ (CS of $\bL_c$) converges to the RS of $\bL$ (CS of $\bL$) in very few iterations.

\subsection{Solving the convex optimization problems} 
The proposed methods are based on convex optimization problems, which can be solved using
generic convex solvers. However, the generic solvers do not scale well to high dimensional data.  In this section, we use an Alternating Direction Method of Multipliers (ADMM) method \cite{boyd2011distributedn} to develop an efficient algorithm for solving (\ref{eq:D_s}). The optimization problem (\ref{eq:mainopt}) can be solved using this algorithm as well, where we would just need to substitute $\bD_r$ and $\bD_r^s$ with $\bD$. 

The optimization problem (\ref{eq:D_s}) can be rewritten as 
\begin{eqnarray}
\begin{aligned}
& \underset{\bQ , \bA , \bB}{\min}
& & \| \bQ \|_1 + \gamma \| \bB^T \|_{1,2} \\
& \text{subject to}
& & \bA = \bB \quad \text{and} \quad \bQ = \bD_r - \bD_r^s \bA  \: , 
\end{aligned}
\label{dokht}
\end{eqnarray}
which is equivalent to
\begin{eqnarray}
\begin{aligned}
 &\underset{\bQ , \bA , \bB}{\min} \: \:
 \| \bQ \|_1 + \gamma \| \bB^T \|_{1,2} + \frac{\mu}{2} \| \bA - \bB \|_F^2 \\ 
& \quad \quad \quad \quad \quad \quad + \frac{\mu}{2} \| \bQ - \bD_r + \bD_r^s \bA \|_F^2 \\
 & \text{subject to} \quad 
 \bA = \bB \quad \text{and} \quad \bQ = \bD_r - \bD_r^s \bA  \: , 
\end{aligned}
\label{dokht2}
\end{eqnarray}
where $\mu$ is the tuning parameter. 
The  Lagrangian function of (\ref{dokht2}) can be written as 
\begin{eqnarray}
\begin{aligned}
 & \calL(\bA,\bB,\bQ,\bY_1,\bY_2) =  
 \| \bQ \|_1 + \gamma \| \bB^T \|_{1,2} + \frac{\mu}{2} \| \bA - \bB \|_F^2 \\
& \quad \quad + \frac{\mu}{2} \| \bQ - \bD_r + \bD_r^s \bA \|_F^2 +  \text{tr} \left( \bY_1^T (\bB-\bA)\right) \\
& \quad \quad + \text{tr} \left( \bY_2^T (\bQ - \bD_r + \bD_r^s \bA)\right)  \: ,\nonumber 
\end{aligned}
\label{dokht3}
\end{eqnarray}
where 
$\bY_1$ and $\bY_2$ are the Lagrange multipliers
and tr$(\cdot)$ denotes the trace of a given matrix.

The ADMM approach then consists of an iterative procedure. Define $\left( \bA_k,\bB_k,\bQ_k, \right)$
as the optimization variables and
$\left( \bY_1^k,\bY_2^k \right)$ as the Lagrange multipliers at the $k^{\text{th}}$ iteration. Define $\bG = \mu^{-1} ( \bI + {\bD_r^s}^T \bD_r^s )^{-1}$ and define the element-wise function $\calT_{\epsilon}(x)$ as $\calT_{\epsilon}(x) = \sgn(x) \max( |x| - \epsilon , 0)$. In addition, define a column-wise thresholding operator $\bZ = \calC_{\epsilon} (\bX)$ as: set
$\bz_i$ equal to zero  if $\| \bx_i\|_2 \leq \epsilon$, otherwise set $\bz_i = \bx_i - \epsilon \: \bx_i/ \|\bx_i\|_2 $, where $\bz_i$ and $\bx_i$ are the $i^{\text{th}}$ columns of $\bZ$ and $\bX$, respectively. 
Each iteration consists of the following steps:\\
1. 
Obtain $\bA_{k+1}$ by minimizing the Lagrangian function with respect to $\bA$
while the other variables are held constant. The optimal $\bA$ is
obtained as
\begin{eqnarray}
\bA_{k+1} = \bG \left( \mu \bB_k + \mu {\bD_r^s}^T(\bD_r - \bQ_k) + \bY_1^k -{\bD_r^s}^T\bY_2^k   )\right)\nonumber
\end{eqnarray}
2. Similarly, $\bQ$ is updated as
\begin{eqnarray}
\bQ_{k+1} = \calT_{\mu^{-1}} \left( \bD_r - \bD_r^s \bA_{k+1} - \mu^{-1} \bY_2^k   \right) \: .\nonumber
\end{eqnarray}

\noindent
3. Update $\bB$ as 
\begin{eqnarray}
\bB_{k+1} = \calC_{\gamma \mu^{-1}} \left( \bA_{k+1} - \mu^{-1} \bY_1^k \right)\: .\nonumber
\end{eqnarray}

\noindent
4. Update the Lagrange multipliers as follows
\begin{eqnarray}
\begin{aligned}
& \bY_1^{k+1} = \bY_1^{k} + \mu (\bB^{k+1} - \bA^{k+1})\\
& \bY_2^{k+1} = \bY_2^{k} + \mu (\bQ^{k+1} - \bD_r + \bD_r^s \bA^{k+1}) \:.\nonumber
\end{aligned}
\end{eqnarray}
These 4 steps are repeated until the algorithm converges or the
number of iterations exceeds a predefined threshold. In our numerical experiments, we initialize $\bY_1$, $\bY_2$, $\bB$, and $\bQ$ with zero matrices.

\section{Robustness to Outlying Data Points}
\label{sec:bothcorruption}
In many application, the data contain outlying data points \cite{rahmani2016coherence,lamport15}.
In this section, we extend the proposed sampling algorithm (\ref{eq:mainopt}) to
make it robust to both sparse corruption and outlying data points.  Suppose the given data can be expressed as
\begin{eqnarray}
\bD = \bL + \bC + \bS\:,
\label{eq: extended model}
\end{eqnarray}
where $\bL$ and $\bS$ follow Data model 1. The matrix $\bC$ has some non-zero columns modeling the outliers. The outlying columns do not lie in the column space of $\bL$ and they cannot be decomposed into columns of $\bL$ plus sparse corruption. Below, we provide two scenarios motivating the model (\ref{eq: extended model}).  

I. Facial images with different illuminations were shown to lie in a low dimensional subspace \cite{basri2003lambertian}. Now suppose we have a dataset consisting of some sparsely corrupted face images along with few images of random objects (e.g., building, cars, cities, ...). The images from random objects cannot be modeled as face images with sparse corruption. We seek a sampling algorithm which can find informative face images while ignoring the presence of the random images to identify the different human subjects in the dataset.

II. A users rating matrix in recommender systems can be modeled as a LR matrix owing to the similarity
between people's preferences for different products.  To account for natural variability in user profiles, the LR plus sparse matrix model can better model the data. However, profile-injection attacks, captured by the matrix $\bC$, may introduce outliers in the user rating databases to promote or suppress certain products.
The model (\ref{eq: extended model}) captures both element-wise and column-wise abnormal ratings.

The objective is to develop a column sampling algorithm which is simultaneously robust to sparse corruption and outlying columns. To this end, we propose the following optimization problem extending (\ref{eq:mainopt})
\begin{eqnarray}
\begin{aligned}
& \underset{{\bA}}{\min} \: \:  \| \bD - \bD {\bA} + \bE \|_1  + \gamma \| \bA^T \|_{1,2} + \lambda \| \bE \|_{1,2}   \\
& \quad\quad\quad\quad\quad\text{subject to} \quad  \text{diag}(\bA) = 0 \:.
\end{aligned}
\label{eq:exted_alg}
\end{eqnarray}

\noindent
The matrix $\bE$ cancels out the effect of the outlying columns in the residual matrix $\bD - \bD {\bA}$. Thus, the regularization term corresponding to $\bE$ uses an $\ell_{1,2}$-norm which promotes column sparsity. Since the outlying columns do not follow low dimensional structures, an outlier cannot be obtained as a linear combination of few data columns. The constraint plays an important role as it prevents the scenario where an outlying column is sampled by $\bA$ to cancel itself.  

\begin{figure}[h]
    \includegraphics[width=0.45\textwidth]{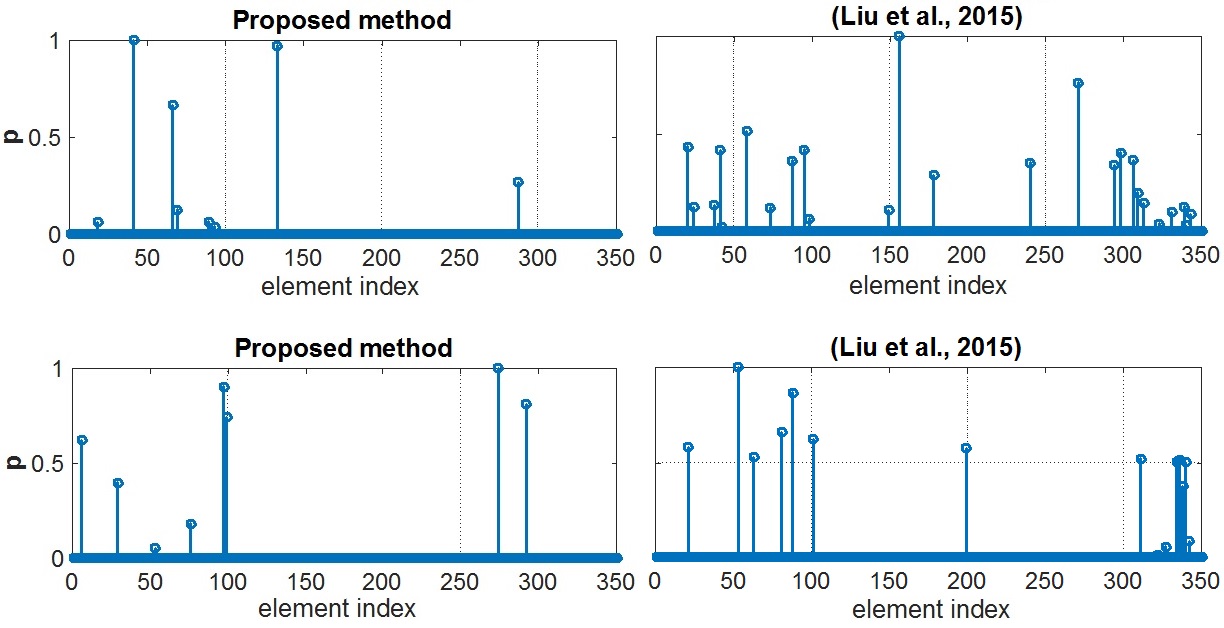}
    \centering
     \vspace{-0.05in}
    \caption{Comparing the performance of (\ref{eq:exted_alg}) with the algorithm in \cite{liu2015robustv,nie2010efficient}. In the first row, $\bD = \bL+\bC+\bS$. In the second row, $\bD = \bL+\bC$. The last 50 columns of $\bD$ are the outliers. }
    \label{fig:2t2}
\end{figure}

The sampling algorithm (\ref{eq:exted_alg}) can locate the representative columns in presence of sparse corruption and outlying columns.
 For instance, suppose $N_1 = 50$, $\bL = [\bL_1 \: \: \bL_2]$, where $\bL_1 \in \mathbb{R}^{50 \times 100}$ and $\bL_2 \in \mathbb{R}^{50 \times 250}$. The ranks of $\bL_1$ and $\bL_2$ are equal to 5 and 2, respectively, and their column spaces are independent. The matrix $\bS$ follows Data model 1 with $\rho = 0.01$ and the last 50 columns of $\bC$ are non-zero. The elements of the last 50 columns of $\bC$ are sampled independently from a zero mean normal distribution. Fig. \ref{fig:2t2} compares the output of (\ref{eq:exted_alg}) with the state-of-the-art robust sampling algorithms in \cite{liu2015robustv,nie2010efficient}. In the first row of Fig. \ref{fig:2t2}, $\bD = \bL+\bS+\bC$, and in the second row, $\bD = \bL+\bC$. Interestingly, even if $\bS = 0$, the robust column sampling algorithm (\ref{eq:exted_alg}) substantially outperforms \cite{liu2015robustv,nie2010efficient}. As shown, the proposed method samples correctly from each cluster (at least 5 columns from the first cluster and at least 2 columns from the second), and unlike \cite{liu2015robustv,nie2010efficient}, does not sample from the outliers. 

\begin{remark}
The sampling algorithm (\ref{eq:exted_alg}) can be used as a robust PCA algorithm. The sampling algorithm is applied to the data and the decomposition algorithm (\ref{eq2n}) is applied to the sampled columns to learn the CS of $\bL$. In \cite{rahmani2016both}, we investigate this problem in more details.  
\end{remark}

\begin{figure}[h]
    \includegraphics[width=0.45\textwidth]{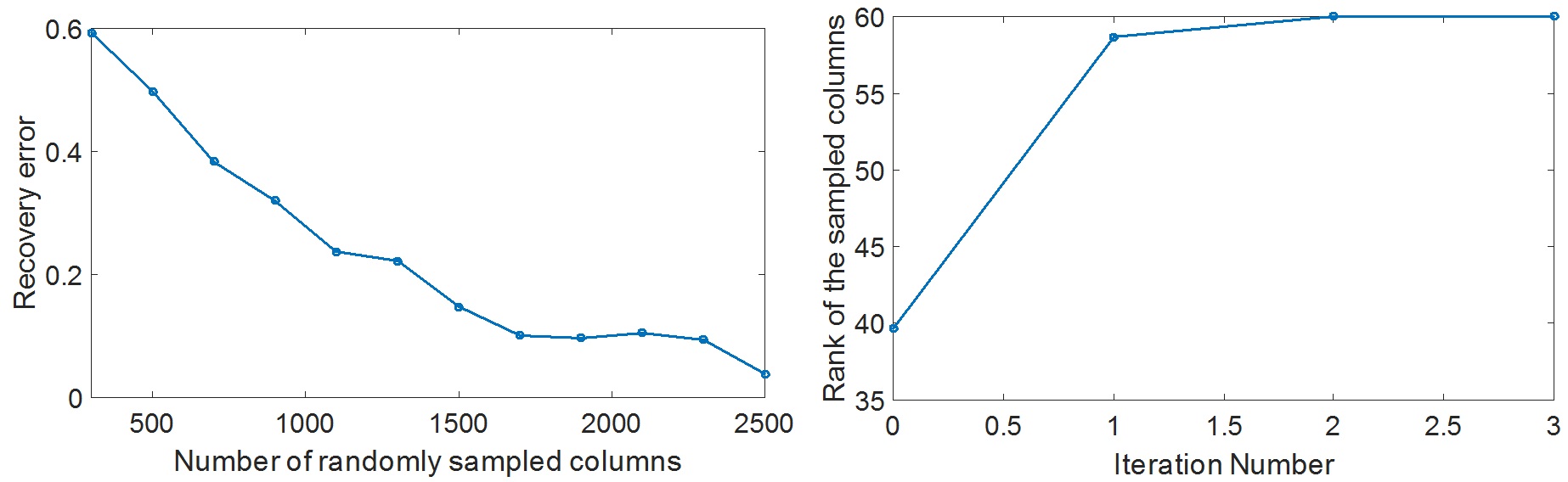}
    \centering
     \vspace{-0.05in}
    \caption{\textbf{Left}: Subspace recovery error versus the number of randomly sampled columns. \textbf{Right}: The rank of sampled columns through the iterations of Algorithm 1. }
    \label{fig:simul1}
\end{figure}

\begin{figure*}[h]
    \includegraphics[width=1.01\textwidth]{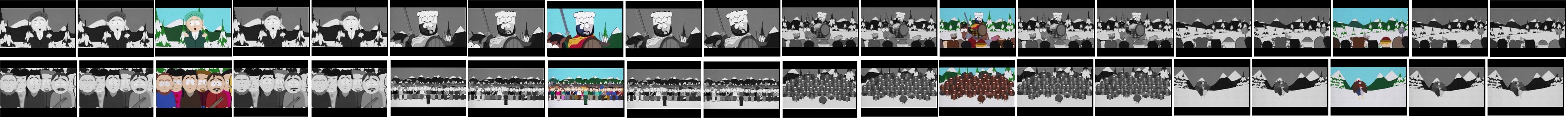}
    \centering
     \vspace{-0.05in}
    \caption{ Few frames of the video file. The frames in color are sampled by the robust column sampling algorithm.  }
    \label{fig:hh1vid}
\end{figure*}

\section{Experimental Results}
In this section, we apply the proposed sampling methods to both real and synthetic data and the performance is compared to the-state-of-the-art.


\subsection{Shortcomings of random sampling}
\label{sec:sim_short}
In the first experiment, it is shown that random sampling uses too many columns from $\bD$ to correctly learn the CS of $\bL$. Suppose the given  data follows Data model 1 with $\rho = 0.01$.  In this experiment,  $\bD$ is a $2000\times6300$ matrix. The LR component is generated as $\bL = [ \bL_1  \: ... \: \bL_{60} ] \: . $
For $1 \leq i \leq 30$, $\bL_i = \bU_i \bQ_i\: ,$
where $\bU_i \in \mathbb{R}^{2000 \times 1}$, $\bQ_i \in \mathbb{R}^{1 \times 200}$ and the elements of $\bU_i$ and $\bQ_i$ are sampled independently from  $\calN(0 , 1)$. For $31 \leq i \leq 60$, $\bL_i = \sqrt{10} \bU_i \bQ_i\: ,$
where $\bU_i \in \mathbb{R}^{2000 \times 1}$, $\bQ_i \in \mathbb{R}^{1 \times 10}$. Thus, with high probability the columns of $\bL$ lie in a union of 60 independent 1-dimensional linear subspaces.

We apply Algorithm 1 to $\bD$.
 The matrix $\bD_r$ is formed using 100 randomly sampled rows of $\bD$.
Since the rows of $\bL$ do not follow a clustering structure, the rank of $\bL_r$ (the LR component of $\bD_r$) is equal to 60 with overwhelming probability.
 The right plot of Fig. \ref{fig:simul1}  shows the rank of the LR component of sampled columns (the rank of $\bL_r^s$) after each iteration of \textbf{Algorithm 1}. Each point is obtained by averaging over 20 independent runs. One can observe that 2 iterations suffice to locate the descriptive columns. We run \textbf{Algorithm 1} with $k_{\max} = 3$. It samples 255 columns on average.

We apply the decomposition algorithm to the sampled columns. Define the recovery error as $\| \bL - \hat{\bU} \hat{\bU}^T \bL \|_F / \| \bL \|_F$, where $\hat{\bU}$ is the basis for the CS learned by decomposing the sampled columns. If we learn the CS using the columns sampled by \textbf{Algorithm 1}, the recovery error in 0.02 on average. On the other hand, the left plot of Fig. \ref{fig:simul1}
 shows the recovery error based on columns sampled using uniform random sampling. As shown, if we sample 2000 columns randomly, we cannot outperform the performance enabled by the informative column sampling algorithm (which here samples 255 columns on average).

\subsection{Face sampling and video summarization with corrupted data}

\begin{figure}[h]
    \includegraphics[width=0.45\textwidth]{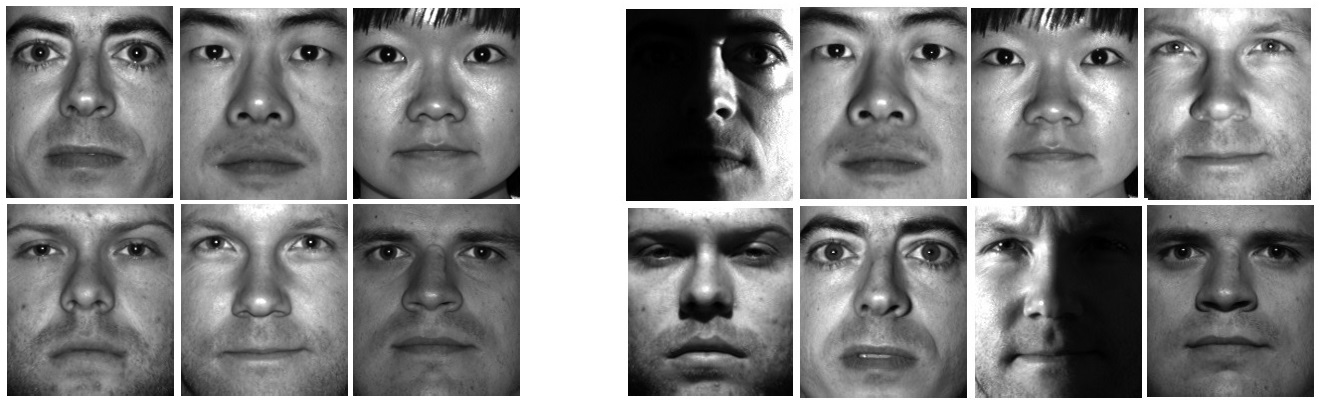}
    \centering
     \vspace{-0.02in}
    \caption{ \textbf{Left}: 6 images from the 6 human subjects forming the dataset. \textbf{Right}: 8 images corresponding to the columns sampled from the face data matrix.}
    \label{fig: faces}
\end{figure}

In this experiment, we use the face images in the Yale Face Database B \cite{lee2005acquiring}. This dataset consists of face images from 38 human subjects. For each subject, there is 64 images with different illuminations. We construct a data matrix with images from 6 human subjects (384 images in total with $\bD \in \mathbb{R}^{32256 \times 384}$ containing the vectorized images). The left panel of Fig. \ref{fig: faces} shows the selected subjects. It has been observed that the images in the Yale dataset follow the LR plus sparse matrix model \cite{lamport2}. In addition, we
randomly replace 2 percent of the pixels of each image with random pixel values, i.e., we add a synthetic sparse matrix (with $\rho = 0.02$) to the images to increase the corruption. The sampling algorithm (\ref{eq:mainopt}) is then applied to $\bD$.  The right panel of Fig. \ref{fig: faces} displays the images corresponding to the sampled columns. Clearly, the algorithm chooses at least one image from each subject.

Informative column sampling algorithms can be utilized for video summarization \cite{elhamifar2014dissimilarity,elhamifar2012see}.
In this experiment, we cut 1500 consecutive frames of a cartoon movie, each with $320 \times 432$ resolution.
The data matrix is formed by adding a sparse matrix with $\rho = 0.02$ to the vectorized frames. 
The sampling algorithm (\ref{eq:semif}) is then applied to find the representative columns (frames).
The matrix $\bD_w$ is constructed using 5000 randomly sampled rows of the data.
The algorithm samples 52 frames which represent almost all the important instances of the video. Fig. \ref{fig:hh1vid} shows some of the sampled frames (designated in color) along with neighboring frames in the video. 
The algorithm judiciously samples one frame from frames that are highly similar.

\subsection{Sampling from highly structured data}

Suppose $\bL_1 \in \mathbb{R}^{2000 \times 6300}$ and $\bL_2 \in \mathbb{R}^{2000 \times 6300}$ are generated independently similar to the way $\bL$ was constructed in Section \ref{sec:sim_short}. Set $\bV$ equal to the first 60  right singular vectors of $\bL_1$, $\bU$ equal to the first 60 right singular vectors of $\bL_2$, and set $\bL = \bU \bV^T$. Thus, the columns/rows of $\bL \in \mathbb{R}^{6300 \time 6300}$ lie in a union of 60 1-dimensional subspaces and their distribution is highly non-uniform. Define $r_w$ and $r_c$ as the rank of $\bL_w$ and $\bL_c$, respectively. Table \ref{tab:rank} shows $r_w$ and $r_c$ through the iterations of \textbf{Algorithm 2}. The values are obtained as the average of 10 independent runs with each average value fixed to the nearest integer.
Initially, $r_w$ is equal to 34. Thus, \textbf{Algorithm 1} is not applicable in this case since the rows of the initial $\bL_w$ do not span the RS of $\bL$. According to Table \ref{tab:rank}, the rank of the sampled columns/rows increases through the iterations and converge to the rank of $\bL$ in 3 iterations.

\begin{table}[h]
\centering
\caption{Rank of columns/rows sampled by \textbf{Algorithm 2}}
\begin{tabular}{| c | c | c | c | c |}
\hline
Iteration Number & 0 &     1 & 2 & 3  \\
 \hline
 $r_c$ & - & 45  & 55  &  60\\
    \hline
 $ r_w $ & 34 & 50 & 58  & 60 \\
  \hline
\end{tabular}
\label{tab:rank}
\end{table}

\subsection{Robustness to outliers}
\begin{figure}[h]
    \includegraphics[width=0.5\textwidth]{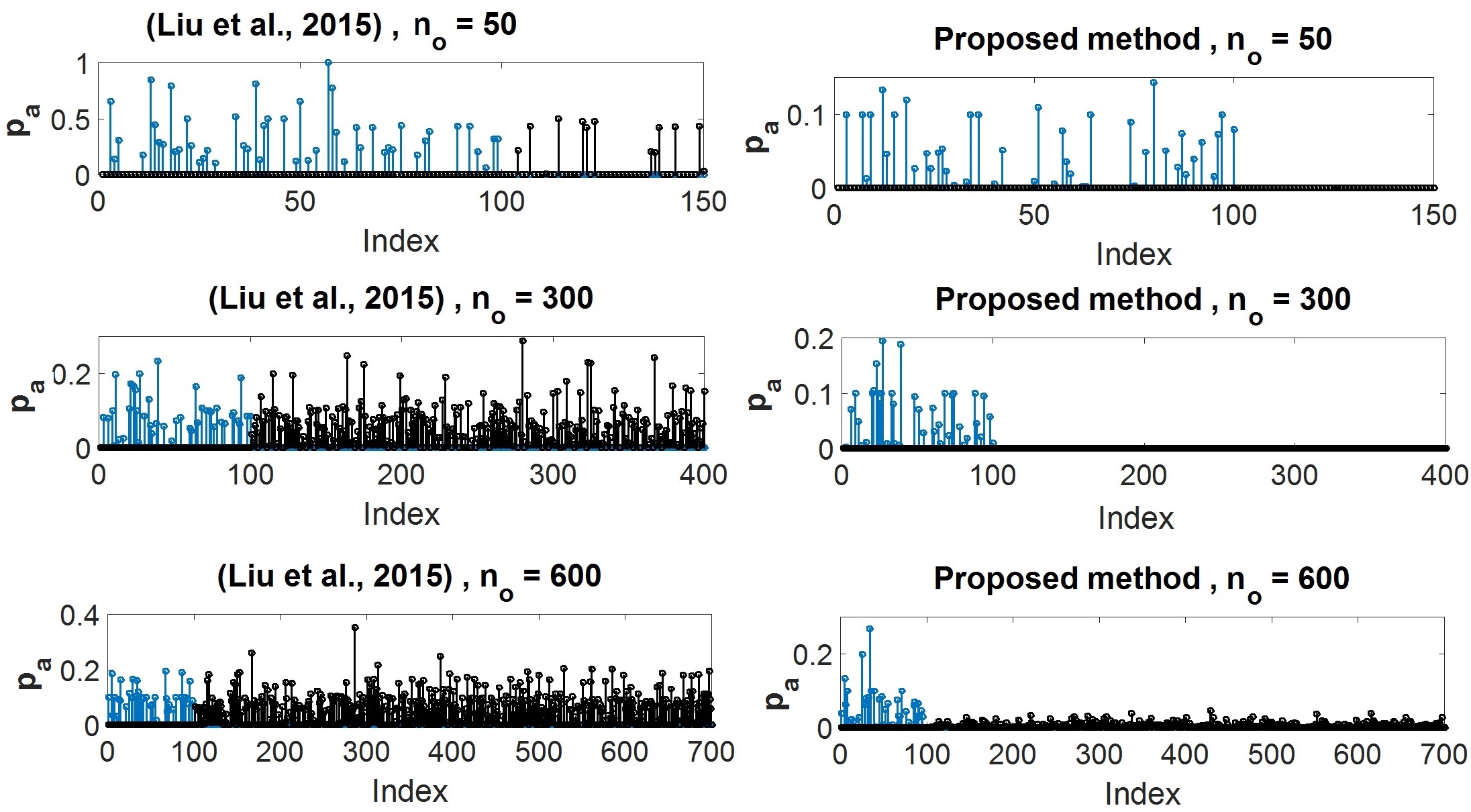}
    \centering
     \vspace{-0.17in}
    \caption{Comparing the performance of (\ref{eq:exted_alg}) with \cite{liu2015robustv,nie2010efficient}.   The blues are the sampled points from inliers and blacks are sampled points from outliers. }
    \label{fig:outll}
\end{figure}

In this experiment, the performance of the sampling algorithm (\ref{eq:exted_alg}) is compared to the state of the art robust sampling algorithm presented in \cite{liu2015robustv,nie2010efficient}.
Since the existing algorithms are not robust to sparse corruption, in this experiment matrix $\bS$ is set equal to zero. The data can be represented as $\bD = [\bD_l \: \: \bD_c]$. The matrix $\bD_l \in \mathbb{R}^{50 \times 100}$ contain the inliers, the rank of $\bD_l$ is equal to 5 and the columns of $\bD_l$ are distributed randomly within the CS of $\bD_l$. The columns of $\bD_o \in \mathbb{R}^{50 \times n_o}$ are the outliers, the elements of $\bD_c$ are sampled from $\calN (0,1)$ and $n_o$ is the number of outliers.  We perform 10 independent runs. Define $\bp_a$ as the average of the 10 vectors $\bp$. Fig. \ref{fig:outll} compares the output of the algorithms for different number of outliers. One can observe that the existing algorithm fails not to sample from outliers. Interestingly, eve if $n_o = 300$, the proposed method does not sample from the outliers. However, if we increase $n_o$ to 600, the proposed method starts to sample from the outliers.

\newpage
{
\bibliographystyle{ieee}
\bibliography{bibfile}
}

\end{document}